\documentclass[conference]{IEEEtran}
\usepackage{times}

\usepackage[numbers]{natbib}
\usepackage{multicol}
\usepackage[bookmarks=true]{hyperref}
\usepackage{cite}
\usepackage{amsmath,amssymb,amsfonts}
\usepackage{algorithmic}
\usepackage{caption}
\usepackage{ulem}
\usepackage{subcaption}
\usepackage{graphicx}
\usepackage{textcomp}
\usepackage{xcolor}
\usepackage{float}
\usepackage{todonotes}
\usepackage{xcolor}
\usepackage{bm}
\usepackage{amsmath}
\usepackage{stfloats}
\renewcommand{\vec}[1]{\bm{#1}}
\newcommand{\mat}[1]{\bold{#1}}

\begin{document}

\title{Precise Object Sliding with Top Contact via Asymmetric Dual Limit Surfaces}


\author{\authorblockN{Xili Yi}
\authorblockA{Department of Robotics\\
University of Michigan\\
Michigan, Ann Arbor 48109\\
Email: yixili@umich.edu}
\and
\authorblockN{Nima Fazeli}
\authorblockA{Department of Robotics\\
University of Michigan\\
Michigan, Ann Arbor 48109\\
Email: nfz@umich.edu}}

\maketitle

\begin{abstract}
In this paper, we discuss the mechanics and planning algorithms to slide an object on a horizontal planar surface via frictional patch contact made with its top surface. 
Here, we propose an asymmetric dual limit surface model to determine slip boundary conditions for both the top and bottom contact. With this model, we obtain a range of twists that can keep the object in sticking contact with the robot end-effector while slipping on the supporting plane. Based on these constraints, we derive a planning algorithm to slide objects with only top contact to arbitrary goal poses without slippage between end effector and the object. We validate the proposed model empirically and demonstrate its predictive accuracy on a variety of object geometries and motions. We also evaluate the planning algorithm over a variety of objects and goals demonstrate an orientation error improvement of 90\% when compared to methods naive to linear path planners.
For more results and information, please visit \url{https://www.mmintlab.com/dual-limit-surfaces/}.
\end{abstract}

\IEEEpeerreviewmaketitle

\section{Introduction}

\begin{figure} 
\centering
\includegraphics[width=8.5cm]{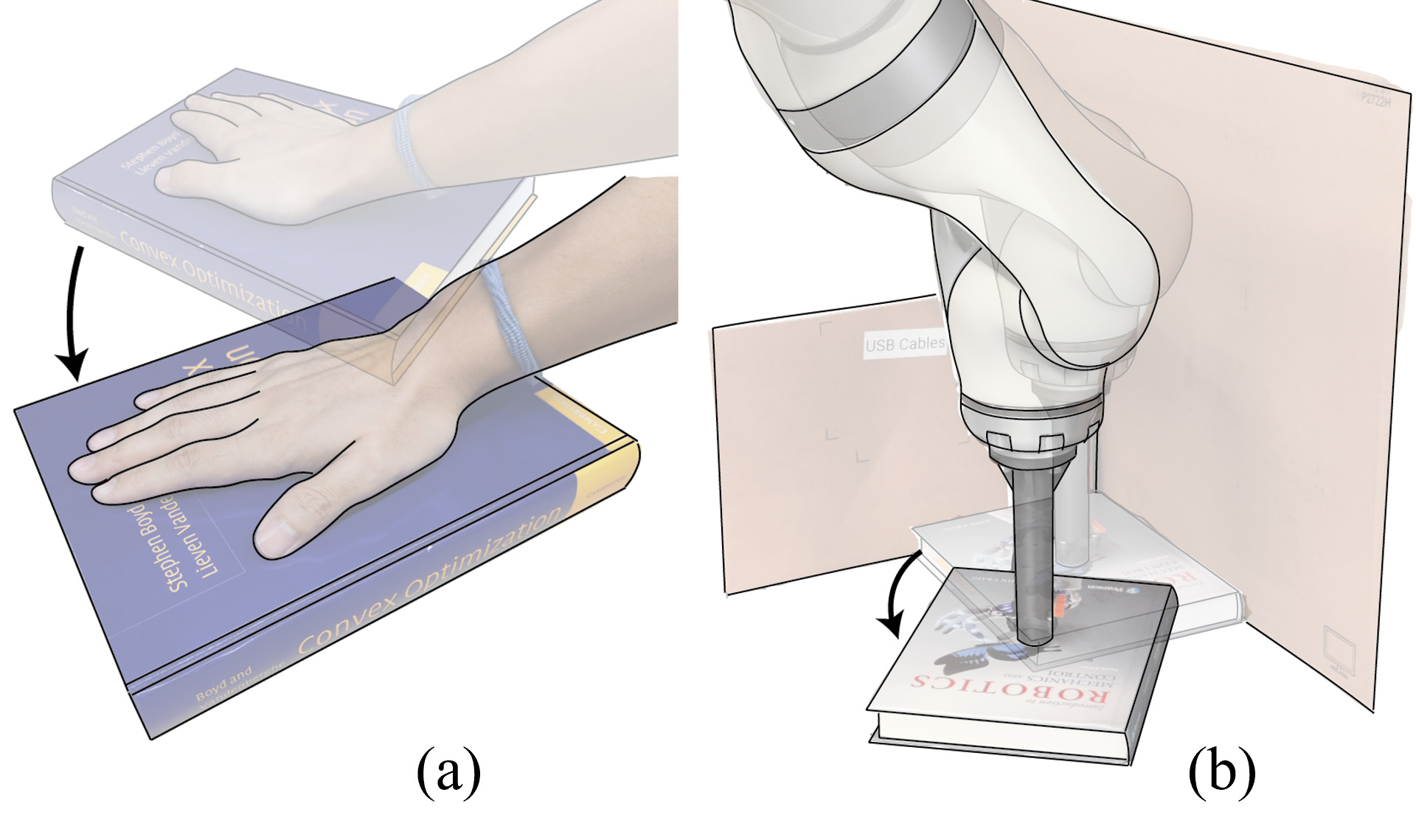}
\caption{(a) Human sliding a book with palm on a horizontal surface. (b) Robot sliding a book with top contact. Here, the book is trapped by two walls meaning that planar pushing cannot solve this task due to side occlusion.}  
\label{fig:teaser_1}
\end{figure}

Planar sliding is an important skill allowing robots to move objects that are too large to fit in standard grippers, too heavy to lift, or are occluded such that grasps are not possible \citep{9811686}. Planar pushing, referring to when the robot pushes on the exposed sides of an object, has recently garnered significant attention \citep{stuber2020let,yu2016more,agrawal2016learning,li2018push,suresh2021tactile,byravan2017se3}. However, planar pushing is not possible when the sides of the object are occluded. For example, Fig.~\ref{fig:teaser_1} shows that the robot is unable extract the book from the corner by pushing on its sides. However, this task is possible by imparting a wrench through the frictional contact patch made on the book's top surface.

In this paper, we propose an asymmetric dual limit surface model that governs the motion of planar objects subject to patch frictional contacts on both the top and bottom surfaces. This model characterizes stick/slip boundaries at both surfaces as a function of the coefficients of friction, object inertia, contact patch sizes, and applied normal forces. We additional propose an open-loop stable planning algorithm that exploits this model to compute robot trajectories that maintain sticking contact with the object while sliding it on the surface. The plans are easily deployed with a simple impedance controller and enable precise sliding of object to arbitrary goal poses.

Our work is closest to \citep{ghazaei2020quasi}, where a pivoting planner and controller were developed to sliding an object to arbitrary goal \textit{orientations}. The robot end-effector only translates and is allowed to slip w.r.t. to the object. While effective, the proposed strategy is focused on object orientation control and requires accurate force feedback. In contrast, our method produces sticking contacts between the object and robot to achieve desired poses, is open-loop stable, and only requires a simple vertical impedance controller to maintain normal force within generous bounds. Also related, \citep{9811686} proposes a friction cone based approach to sliding with pure translation but does not control object orientation.

\subsection{Problem statement}
Consider the exemplar sliding task shown in Fig. \ref{fig:teaser_1}(b), where the robot is to slide the object on a horizontal surface to a desired pose.
We assume that the robot has established a patch contact above the geometric center of the object and is applying normal force \(N_e(t)\) to the top surface. Further, we assume quasi-static motion (inertial forces are negligable) and that the end-effector, object, and support are rigid. 

\begin{figure} 
\centering
\includegraphics[width=7cm]{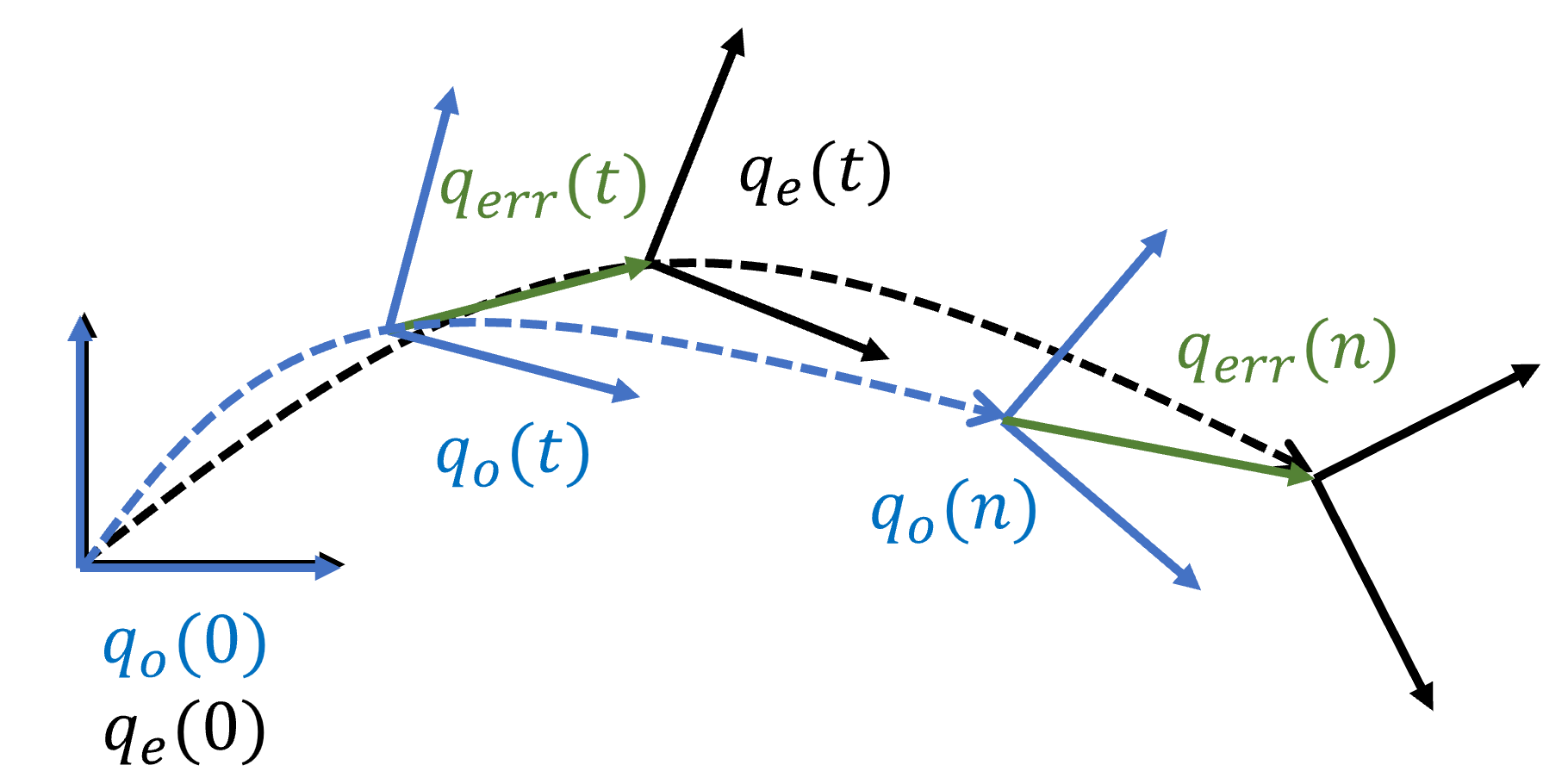}
\caption{Schematic of the sliding problem. 
Here, \(\vec{q_e}(t)\), \(\vec{q_o}(t)\)$\in$ \(\mathbb{R}^3\) are the 2D pose of the end-effector (black frames) and the object (blue frames) respectively. At time \(t=0\), \(\vec{q_e}(0)\), \(\vec{q_o}(0)\) are set to be \(\vec{0}\). The green arrows represent the pose error \(q_{err}\) between end-effector and the object.}  
\label{fig:problem_statement}
\end{figure}

Let $\vec{q}_e(t) \in \mathrm{SE}(2)$ 
and $\vec{q}_o(t) \in \mathrm{SE}(2)$ denote the 2D poses of the end-effector and the object, respectively. We define  \(\vec{q}_{err}\) $\in$ \(\mathrm{SE}(2)\) as the relative 2D pose of the object w.r.t. to the robot, i.e., \(\vec{q}_{err}(t)=\vec{q}_e(t) - \vec{q}_o(t)\). As shown in Fig.~\ref{fig:problem_statement}, at the beginning of each path, the end effector is aligned with the object, i.e., \( \vec{q}_{err}(0) = 0\). 
Then the relative 2D pose along the path \(\vec{q}_{err}(t)\) only deviates from its initial value due to slip between the end-effector and object. Given an arbitrary object goal pose \(\vec{q}_g \in \mathrm{SE}(2)\), our goal is to design an input composed of the end-effector trajectory and normal force \(\vec{u}(t)=[\vec{q}_e(t), N_e(t)]\) 
to minimize the final pose error:
\begin{equation*}
\vec{u}^* = \arg \min_{\vec{u}} ||\vec{q}_{o}(T) - \vec{q}_g ||\label{eq:ctrlprob}
\end{equation*}
where \(T\) is time at the end of the path. Our key insight is to compute $\vec{u}(t)$ such that $\vec{q}_{err}(t)=0$ along the entire path and $\vec{q}_e(T) = \vec{q}_g$. Intuitively, the first constraint imposes sticking contact between the robot and object, and the second constraint imposes that the end-effector reaches the goal object pose, driving the object to this pose.



\section{Related work}
Planar manipulation contact mechanics have a long and rich history in robotics. Howe et al. \citep{howe1988sliding} studied the contact between a robot finger and an object during slippage under both torsion and shear. Goyal \citep{goyal1989planar} defined the limit surface as the boundary of the set of all possible friction wrenches to characterize friction between two surfaces. In a subsequent work from Howe et al. \citep{howe1996practical}, different models for the limit surface were proposed and compared, with a simplified ellipsoidal model being proposed for reducing computation complexity. Xydas et al. \citep{xydas1999modeling} presented modeling and experimental results to evaluate the relationship between the normal force and the radius of contact for soft fingers. Shi et al. \citep{shi2017dynamic} derived a model of sliding dynamics based on the limit surface contact model at each fingertip. In this work, we also use the ellipsoidal approximation of the limit surface to model friction behavior for each contact and extend these models to a multi-contact scenario.
    
Modeling, planning and control of planar pushing have been widely studied.
\citet{stuber2020let} conducted an extensive survey of methods of robot pushing planning and control, including both analytical and data-driven methods. Lynch and Mason \citep{lynch1996stable} proposed a motion planning algorithm that considers the sticking interaction to find trajectories towards target goals. \citet{agrawal2016learning} introduced a method for learning the intuitive physics and planning of unknown objects using deep neural networks and visual data obtained through poking. \citet{li2018push} presented Push-Net, a deep recurrent neural network model that enables a robot to reposition and reorient objects. Push-Net learns the center of mass of unknown objects and predicts motion through visual data obtained through pushing. \citet{suresh2021tactile} proposed a method to estimate both object shape and pose in real-time from a stream of tactile measurements. \citet{byravan2017se3} introduced SE3-NETS, deep neural networks designed to model and learn rigid body motion from raw point cloud data. They demonstrated the performance of SE3-NETS in both simulation and real-world scenarios. 
Yu et al. \citep{yu2016more} conducted a comprehensive study of different friction models by pushing objects with various materials, shapes, contact locations, directions, velocities, and accelerations. We choose the material for experiments based on this work.

The problem of planar manipulation with top contacts \citep{kao1992quasistatic,kao1993comparison,xue1994dexterous,ghazaei2020quasi,fakhari2019modeling,9811686,zhou2016convex} is less discussed than planar pushing problems. This type of manipulation is usually called as dragging or sliding, in comparison with pushing, which relies on normal force on object sides. All these works also rely on a quasi-static analysis and Coulomb friction with limit surface modeling, as well as soft contacts that allow moment transfer. 
Kao et al. \citep{kao1992quasistatic} studied a generic manipulation problem with compliance and sliding, followed by studies of the mechanics of manipulating a business card with two sliding robot fingers on a frictionless table \citep{kao1993comparison,xue1994dexterous}. However, frictionless assumptions are not realistic when sliding heavier objects.
The work in \citep{ghazaei2020quasi} derives a hybrid dynamical system, validated via simulations and experiments, to predict and control the motion of the object and the interaction forces, providing conditions for sticking, slipping and pivoting. They also posed a potential planning problem of finding trajectories to slide an object from an initial to a desired pose such that the contact between surfaces is maintained and the normal force is within desired limits, though the problem remains unsolved in the work.
The work in \citep{fakhari2019modeling} discussed the design of a controller to reduce the undesired slippage between top surface of object and robot finger of an unknown object, but this work is performed only in simulation and rely on accurate force/torque control.
The work in \citep{9811686} includes a model free method for dragging an object to manipulate object with unknown dynamic and friction parameters. However, this works focused on planar translation, while not considering controlling rotation by sliding. 
The work in \citep{zhou2016convex} proposed a polynomial force-motion model for planar sliding with data-driven methods.

In the sliding problem we discuss in this paper, we proposed a path planning algorithm to prevent slippage between end-effector and object under a specified normal force while sliding. This slippage-free planning algorithm aim to work not only for translation but also rotation, without force feedback along the path.


\section{Mechanics of planar slippage-free sliding}



To drive the object pose to its desired value with an open-loop controller, our key insight is to choose robot motions such that \(\vec{q_{err}}(t)=\vec{0}\; \forall \; t \), i.e., maintaining sticking contact (no slip) between the object and the end-effector while the object slides on the supporting plane along the entire path. We refer to this motion as slippage-free sliding. To achieve slippage-free sliding, here we derive the asymmetric dual limit surface model that describes resultant object motions subject to patch top and support contacts. 
We first discuss the mechanics of planar object sliding by reviewing the fundamental concept of limit surfaces. Next, we will present the asymmetric dual limit surface and dual limit cone models. Lastly, we will derive slippage-free twists boundary constraints that result in slippage-free motion. 

\subsection{Limit Surfaces Model}
The limit surface is defined as the boundary of the set of all possible frictional wrenches that a supporting contact patch can offer \citep{goyal1989planar}. This boundary also characterizes the set of all possible instantaneous object twists due to the frictional interaction between the object and its support contact. In Fig.~\ref{fig:teaser_1}(b), when we focus on the book, for the contact between the book and supporting plane, the plane is the supporting contact, while for the contact between the book and end-effector, the end-effector is the supporting contact. 

In general, calculating the limit surface in closed-form is impossible; however, the seminal work of \citet{howe1996practical} proposed an ellipsoidal approximation in wrench space that has proven effective in a variety of later studies \citep{lynch1996stable,xydas1999modeling,hogan2020feedback,hogan2020reactive,kloss2022combining,shi2017dynamic}. Here, we also make use of this model. 
%
%
Let $\vec{w}=[f_x, f_y, m_z]^T$ be a friction wrench on the object from the supporting contact in the contact frame. The ellipsoidal limit surface can be expressed as:
\begin{equation*}
    \vec{w}^T\mat{A}\vec{w}=1
    \label{eqn:limit_surface}
\end{equation*}
where $\mat{A}=\text{Diag}\{(a_1N)^{-2}, (a_2N)^{-2}, (a_3N)^{-2}\}$.
Assuming isotropic friction with Coulomb friction model, we have $a_1=a_2=\mu$, where $\mu$ is the friction coefficient between the contact and the object, and $N$ is the normal force at that contact. We can represent arbitrary contact patches using their equivalent radius \(r_o\) \citep{howe1996practical} to generate the ellipsoidal limit surface. Using this radius, we can write the maximum friction torque about the contact normal as $a_3=r_0c\mu$, where $r_0$ is the equivalent radius of the object and $c\in [0,1]$ is a constant corresponding to object geometry. This constant is obtained by integration and for a uniform pressure distribution with the equivalent radius contact, $c \approx 0.6$ \citep{xydas1999modeling, shi2017dynamic}. 
We can now write the relationship $ a_3 = rca_2=rca_1 $ where we note that for same contact geometry, $A\propto N$, i.e., that the size of the limit surface is proportional to the normal force while maintaining the limit surface axes ratios.

As is standard with the ellipsoidal limit surface approximation \citep{fakhari2019modeling,hogan2020feedback,zhou2016convex, 9811686,ghazaei2020quasi}, we assume that the maximum static friction is equal to the kinetic friction under the same normal force. Static friction wrenches are inside or on the limit surface, while dynamic friction wrenches are on the limit surface. When the object slides on the contact plane, the wrench $\vec{w}_c$ lies on the limit surface and the surface normal at this point is the instantaneous object twist direction \citep{howe1996practical}.

\subsection{Asymmetric Dual Limit Surface and Cones Model} \label{subsection:3_b}

To develop the asymmetric dual limit surfaces model, consider Fig. \ref{fig:teaser_1}(b) where the end-effector is in contact with the top surface of the object while the object is also in contact with a supporting plane. Here, the supporting planes are perpendicular to the direction of gravity. For scenarios in which the end-effector is aligned with the object's center, 
we have two center-aligned contact surfaces. For each contact surface, we can plot a limit surface corresponding to the normal force. These limit surfaces will have different geometries due to inertia of the object and relative contact patch sizes, creating an asymmetry that can be exploited for control.

\begin{figure} 
    \centering
    \includegraphics[width=8cm]{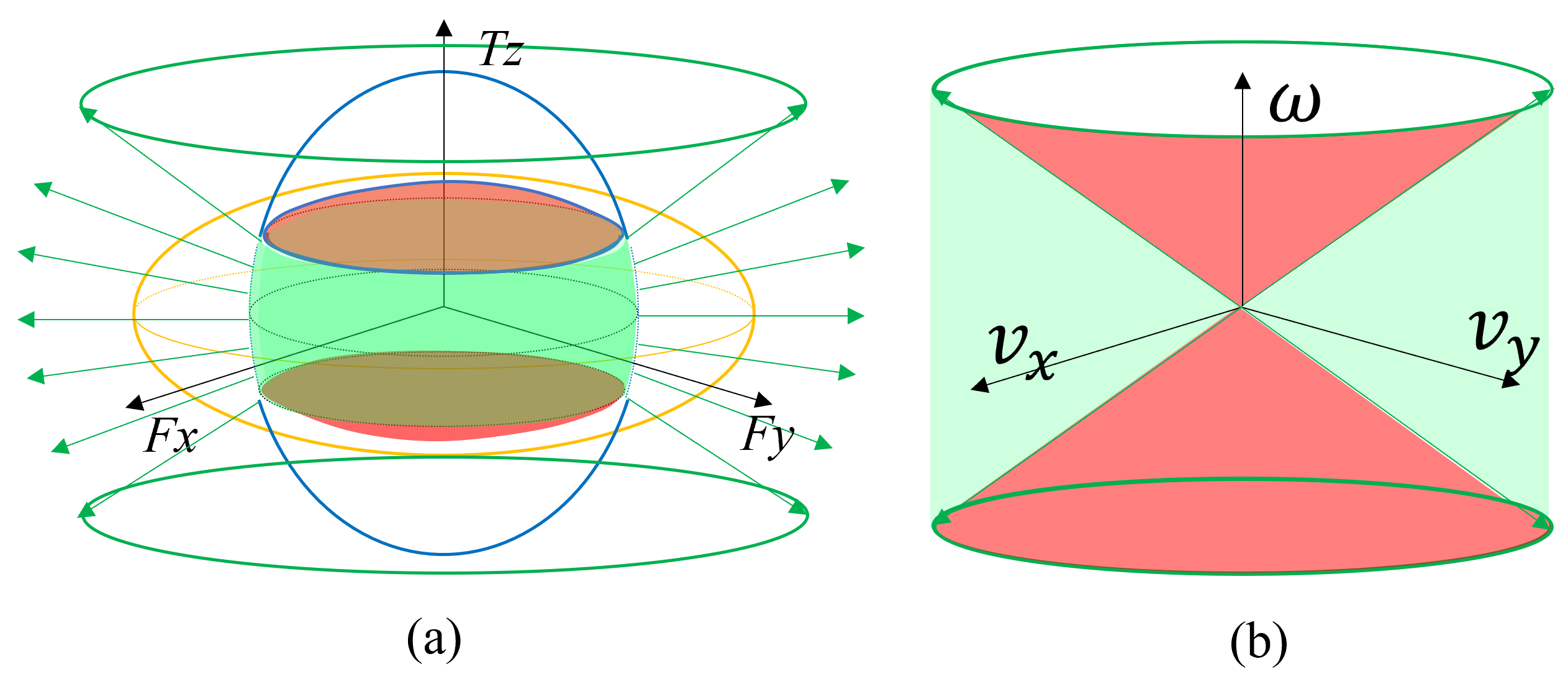}
    \caption{(a) Schematic of dual limit surfaces model in case (III) or (IV) mentioned in Section \ref{subsection:3_b}. The blue ellipse represents the limit surface between the object and end-effector while the orange ellipse represents the limit surface between the object and the supporting plane. Wrenches on the top and bottom red surfaces mean the object is moving on the supporting plane or is static. Wrenches on the green surface mean the object is sticking on the supporting plane while the end-effector slips on the object or is static. Wrenches inside the green and red surfaces mean the object is static. Schematic of slippage-free twists on the dual limit surfaces model in case (III) or (IV) mentioned in Section \ref{subsection:3_b}.  Slippage-free twists are the set of normal directions of all points on the green surface, which is also shown as green volume on panel (c). In cases (II) and (V), the red and green colored regions switch.} 
    \label{fig:slippage-free_twist_with_DLS}
\end{figure}

Let $N_e$ denote the normal force applied by the end-effector to the object, then the normal force between the object and supporting plane \(N_p\) is the sum of the gravitational force and the applied normal force:
\begin{equation*}
    N_p=mg+N_e
    \label{eqn:normal_plane}
\end{equation*}
For a choice of $N_e$ and physical parameters (mass, friction, and equivalence radii), we can plot the 2 limit surfaces in the object wrench space -- Fig. \ref{fig:slippage-free_twist_with_DLS}(a) shows an example. With the quasi-static assumptions, considering force-torque balance in the x-y plane, we have:
\begin{equation*}
\vec{w_e}+\vec{w_p}=0
\label{eqn:balance_plane}
\end{equation*}
where \(\vec{w_e}\) is the wrench applied by the end-effector to the object projected to supporting plane contact frame, \(\vec{w_p}\) is the wrench applied from the supporting plane to the object.

These two wrenches are equal in magnitude but opposite in direction, considering that both limit surfaces are symmetric to the origin point, we can use single point in this dual limit surfaces to represent this pair of friction wrenches.  
\begin{figure*} 
\centering
\includegraphics[width=170mm]{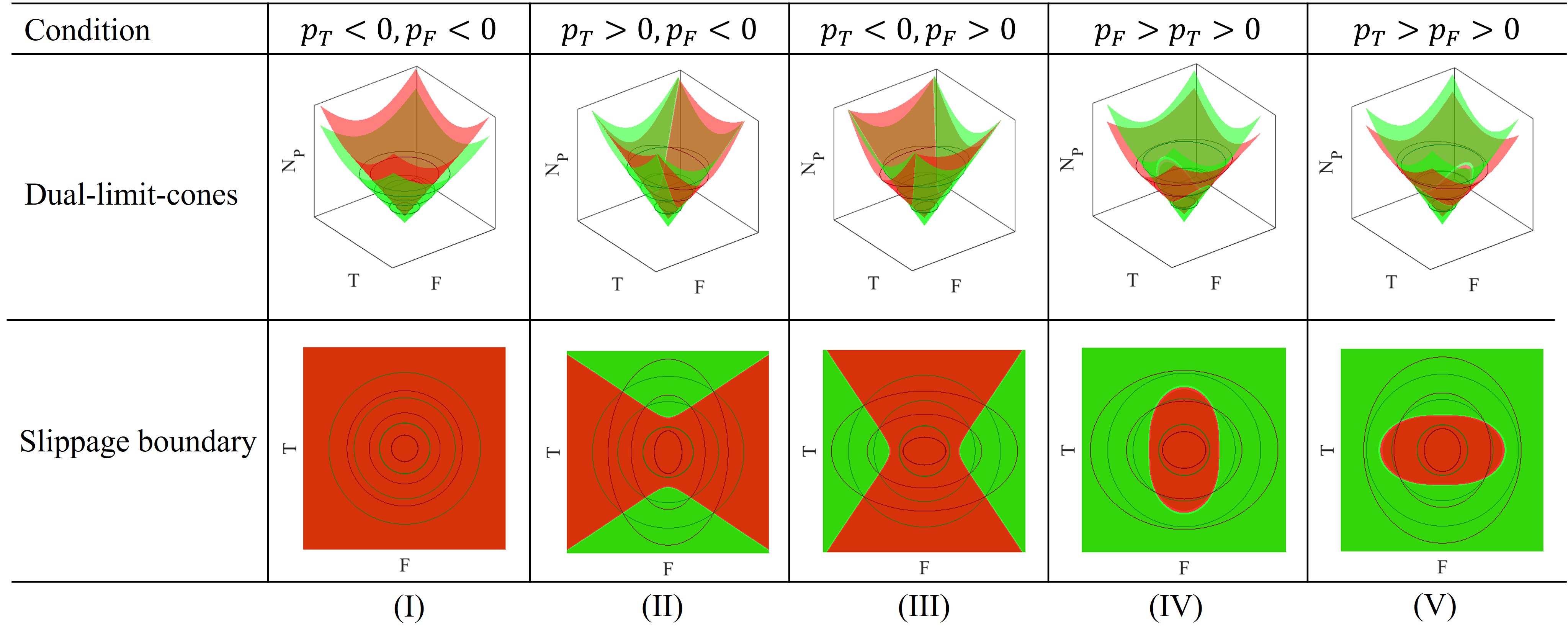}

\caption{Schematic of dual limit cones in 5 cases. The middle row shows dual limit cone schematics, where red surfaces represent the contact between the end-effector and object while the green surfaces represent the contact between object and supporting plane. The bottom row shows the projections of dual limit cones on to the F-T plane. Also, overlayed are the 2D dual limit surfaces shown as ellipses for a range of normal forces. The slippage boundaries are the points at which the ellipses intersect and are the boundaries between the red and green surfaces. 
}
\label{fig:dual_limit_surface_different_cases}
\end{figure*}

Both \(\vec{w_e}\) and \(\vec{w_p}\) must be inside or on their respective limit surfaces. As a result, the frictional wrench should be inside or on the intersection of both limit surfaces in Fig. \ref{fig:slippage-free_twist_with_DLS}(a). 
For each type of 
ellipsoidal limit surfaces interaction, there are at most 3  possible motions: i) neither the object or the end-effector moves (inside both green and red surfaces), ii) the object slips on the supporting plane while sticking to the end-effector (on green surface), and iii) the object sticks to the supporting plane while the end-effector slips on the object (on red surface). 

There are 5 distinct types of ellipsoid interactions, as shown in Fig. 4 (where Fig. 3 is an example of cases (II) and (V)).
To visualize this, we first note that planar translation along the \(F_x\) and \(F_y\) axes are equivalent due to the assumption of symmetric and homogeneous friction properties. Thus, we can reduce the dual limit surfaces at a given normal force to a 2D space where one axis represents the friction force and the other represents friction torque. Next, we add the axis of normal force \(N_p\) to visualize how dual limit surfaces change with respect to the normal force. We call this construct the dual limit cone, visualized in Fig. \ref{fig:dual_limit_surface_different_cases}. This model extends the work in \citet{9811686}, as the cross-section along \(T\)-\(N_p\) plane is the Friction Cone Diagram with pure translation. Cross-sections at a given \(N_p\) parallel to T-F plane represent dual limit surfaces at that normal force. The two cones are shifted along the \(N_p\) axis by the magnitude of the gravitational force acting on the object. The projection of the intersection of the two cone limit surfaces to the F-T plane represents the boundary of slippage for all normal forces.
These five cases are determined by the following parameters:
\begin{equation*}
p_T = ({\mu_e r_e - \mu_p r_p})/({\mu_p r_p}), \quad p_F = ({\mu_e - \mu_p})/{\mu_p}
\end{equation*}
where \(p_T\) and \(p_F\) are the differences between the slopes of the two cones along the \(T\) and \(F\) axis, respectively.
These five cases are shown in five columns in Fig. \ref{fig:dual_limit_surface_different_cases}. In each column, the top row is the condition, the middle is the 3D visualization of the dual limit cones, and the bottom is their projection on the F-T plane. This projection shows the boundary of slippage 
independent of the normal force \(N_e\). The robot is able to maintain sticking contact with the object by imparting wrenches in the green zone.



Changing normal force will change the relative size of these 2 limit surfaces, as it will change the ratio between normal forces applied through each contact: 
\begin{equation*}
    R=\frac{N_e}{N_e+mg}
    \label{eqn:ratio}
\end{equation*}
Increasing normal force increase the ratio $R$, which results in changing the intersection between the 2 limit surfaces. We note that since $N_e$ and $mg$ are both greater than zero, then the upper limit of \(R=1\). 
In case (I), no matter what the normal force is, the limit surface between the object and end-effector is always inside that of the object and supporting plane, thus the end-effector will always slip on the object. In cases (III) and (IV), when the normal force \(N_e\) is smaller than:
\begin{equation*}
    {N_e}_{slip}=\frac{\mu_pmg}{\mu_e-\mu_p}
    \label{eqn:force_slip}
\end{equation*}
will lead to the relationship described by case (I) where the end-effector will always slip on the object.

In case (IV), increasing the normal force may result in the limit surface of the object and supporting plane to lie entirely inside the limit surface of the end-effector and object, which means the object will always follow the end-effector. The boundary normal force will be:
\begin{equation*}
    {N_e}_{stick}=\frac{r_p\mu_pmg}{r_e\mu_e-r_p\mu_p}
    \label{eqn:force_boundary}
\end{equation*}
when the normal force \(N_e\) is bigger than \({N_e}_{stick}\), the object will always stick to the end-effector. Similarly, we obtain the slip condition for case (II) and (V):
\begin{equation*}
{N_e}_{slip}=\frac{r_p\mu_pmg}{r_e\mu_e-r_p\mu_p}
    \label{eqn:force_slip_be}
\end{equation*}
and the sticking boundary force for case (V):
\begin{equation*}
    {N_e}_{stick}=\frac{\mu_pmg}{\mu_e-\mu_p}
    \label{eqn:force_stick_e}
\end{equation*}

In the following section, we will not discuss case (I) as the object cannot move no matter what wrench is applied. We will discuss cases (II) and (III) with normal force greater than \({N_e}_{slip}\), and case (IV) and (V) with normal force between \({N_e}_{slip}\) and \({N_e}_{stick}\), as in these ranges, case (IV) behaves similarly to (III) and case (V) behaves similarly to case (II).

\subsection{Slippage-free Twist Range}

The object will follow the end-effector when the end-effector has sticking contact with the object, while the object is slipping on the supporting plane. To realize this, the wrench imparted to the object by the end-effector must lie inside its respective limit surface, while the wrench imparted to the object by the support must lie on its respective limit surface, i.e., the green surface in Fig. \ref{fig:slippage-free_twist_with_DLS}(a).
These wrenches are also the set of wrenches that can move the object. The limit surface model determines that for each friction wrench, the object twist will be in the direction of the surface normal of the limit surface at that wrench, as shown in Fig. \ref{fig:slippage-free_twist_with_DLS}(a). Then all normal directions of the aforementioned wrench set  result in the set of all object twists in the plane, shown in Fig. \ref{fig:slippage-free_twist_with_DLS}(b). We define the set of all twists for which the object can follow the end-effector as slippage-free twists. Then for all slippage-free twists, the following relationship holds:
\begin{equation}
    k_v(N_e)\cdot\frac{\sqrt{v_x^2+v_y^2}}{|\omega|} 
    \begin{cases}
    \geq 1,             & \text{case (III) and (IV) } \\
    < 1,             & \text{case (II) and (V)}
\end{cases}
    \label{eqn:slippage-free}
\end{equation}
where $k_v(N_e)$ is the slope of the normal to the 2D limit surface of the object and end-effector at the intersection of the limit surfaces when the normal force is \(N_e\).
%
$k_v$ is the maximum ratio (case (III) and (IV)) or the minimum ratio (case (II) and (V)) between angular velocity and linear velocity without slippage between the end effector and object. 
This slippage-free twist relationship (Eq. \ref{eqn:slippage-free}) is the kinematic constraint for moving any object without slippage between the object and end-effector with the normal force \(N_e\) under the quasi-static assumption.


To provide intuition, consider case Fig.~\ref{fig:dual_limit_surface_different_cases}(c). $k_v$ provides the upper bound for object rotation for a given $\delta=(v_x,v_y)$ linear twist and corresponds to the wrenches at the intersection of the green and red surfaces, i.e., the slippage-free boundary. For any choice of wrench in the green surface, the robot will maintain sticking contact while sliding the object; however, the object rotation will be strictly smaller for the same given $\delta$ linear motion. In practice, we can multiply \(k_v\) by a ``safety factor'' to plan more conservative paths that maintain distance to the boundary and ensure no slippage between the end-effector and object will happen. This safety factor relaxes some of the stronger assumptions made earlier and allows for some uncertainty in the model parameters. We discuss the estimation of the model parameters in the Experiments and Results sections.  


\section{Motion Planning for Sliding}

In this section, we derive a planning algorithm that exploits the slippage-free constraint Eq.~\ref{eqn:slippage-free} to compute open-loop stable robot paths that slide the object to arbitrary poses in the plane. This constraint must hold for every time step.  
To simplify planning, we fix the normal force \(N_e\) throughout the path, thus \(k_v\) is also fixed. Due to the use of the safety factor, the precise value of the normal force during execution does not impact the motion or its stability.

To compute an optimal path, we pose the planning problem as a quadratic programming. We denote planned path as \(\vec{\tau}=[\vec{q}_1^T, \vec{q}_2^T, \dots, \vec{q}_n^T]^T\). Here, \(\vec{q}_1\) and \(\vec{q}_n\) are the start and goal poses, respectively. We also denote linear interpolation path from \(\vec{q}_1\) and \(\vec{q}_n\) as \([\hat{\vec{q}_1}^T, \hat{\vec{q}_1}^T, \dots, \hat{\vec{q}_{n}}^T]^T\).
We want to minimize both the distance from the connecting line between the start and goal poses and the second derivative along the whole path. The first component is responsible for accuracy while the latter is for smoothness. Then the path planning problem can be expressed as a quadratic constraint quadratic programming (QCQP) problem as follows:
\begin{equation*}
\begin{aligned}
\min_{\vec{\tau}} \quad &  C_1\sum_{i=1}^{n}{(\vec{q}_i - \hat{\vec{q}_i})^T(\vec{q}_i - \hat{\vec{q}_i})} \\
&+ C_2\sum_{i=3}^{n}{[\vec{q}_{i-2}^T, \vec{q}_{i-1}^T, \vec{q}_{i}^T]\mat{P}[\vec{q}_{i-2}^T, \vec{q}_{i-1}^T, \vec{q}_{i}^T]^T} \\
\end{aligned}
\end{equation*}

\begin{equation*}
\begin{aligned}
\textrm{s.t.} \quad &  \vec{q}_{1}\mbox{ and } \vec{q}_{n} \mbox{ are given} \\
  &[\vec{q}_{i-1}^T, \vec{q}_{i}^T]\mat{K}[\vec{q}_{i-1}^T, \vec{q}_{i}^T]^T
    \begin{cases}
    >0, & \text{case(III), (IV)} \\
    <0, & \text{case(II), (V)} \\
    \end{cases}\\ 
    & \forall i\in[2,\dots,n] & 
\end{aligned}
\label{eq:qcqp_raw}
\end{equation*}
where $C_1$, $C_2$ are weights of distance from linear interpolation and second derivative. Matrix \(\mat{P}\) accounts for the second derivative of the path, and it is constructed as:
\begin{equation*}
    \mat{P} =(\begin{bmatrix} 1 & -2 & 1\end{bmatrix} ^T\begin{bmatrix} 1 & -2 & 1\end{bmatrix} )\otimes \mat{I}
\end{equation*}
where \(\mat{I} \in \mathbb{R}^{3\times3}\) denotes identity matrix. Matrix \(\mat{K}\) in Eq. (\ref{eq:qcqp_raw}) is the velocity constraint for each time step, and it is constructed as follows:
\begin{equation*}
    \mat{K} =\begin{bmatrix} 1 & -1\\ -1 & 1\end{bmatrix} \otimes diag(k_v, k_v, -1) 
\end{equation*}
Once computed, the optimal robot path $\vec{\tau}$ will drive the object to the desired goal pose.

\section{Experiments}
\subsection{Experiment setup} \label{subsection:experiment_setup}
\begin{figure} 
\centering
\includegraphics[width=8cm]{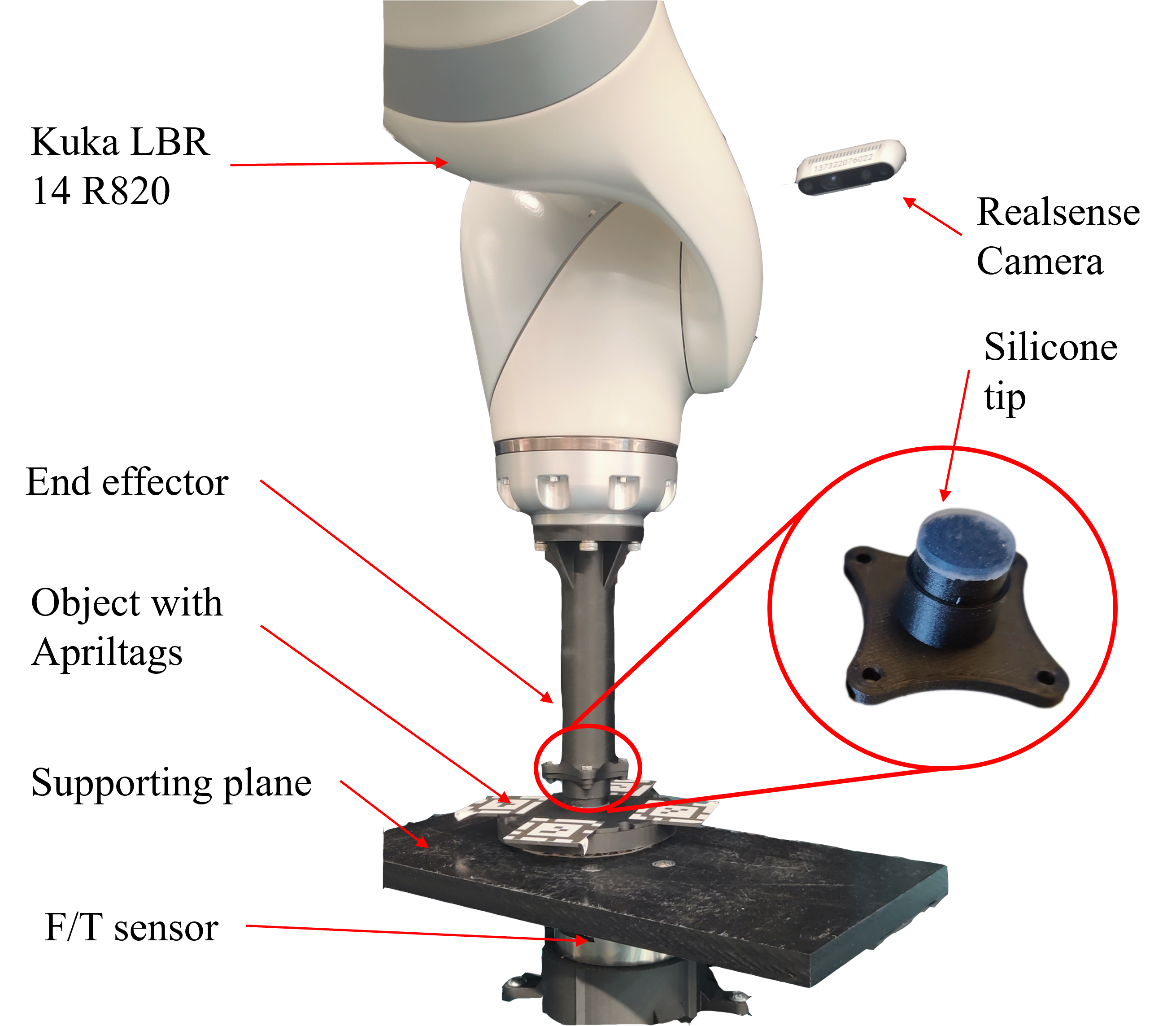}
\caption{Experiment setup for planar sliding with palm manipulation.}
\label{fig:experiment_setup}
\end{figure}

Our experimental setup is shown in Fig. \ref{fig:experiment_setup}. 
Here, the robot is equipped with a 3D-printed end-effector with a silicone gel tip to provide a small amount of compliance to ensure patch contacts. The objects slide on a 0.5 inch thick Derlin/Acetal Copolymer plastic board as the supporting plane mounted on a 6 DOF ATI Gamma force torque sensor (resolution: (0.0125 [N], 0.0007 [Nm])) at a maximum sample-rate of 7000 [Hz]. The board provides a uniformly distributed friction coefficient and has high rigidity. 

\begin{figure} 
    \centering
    \includegraphics[width=8cm]{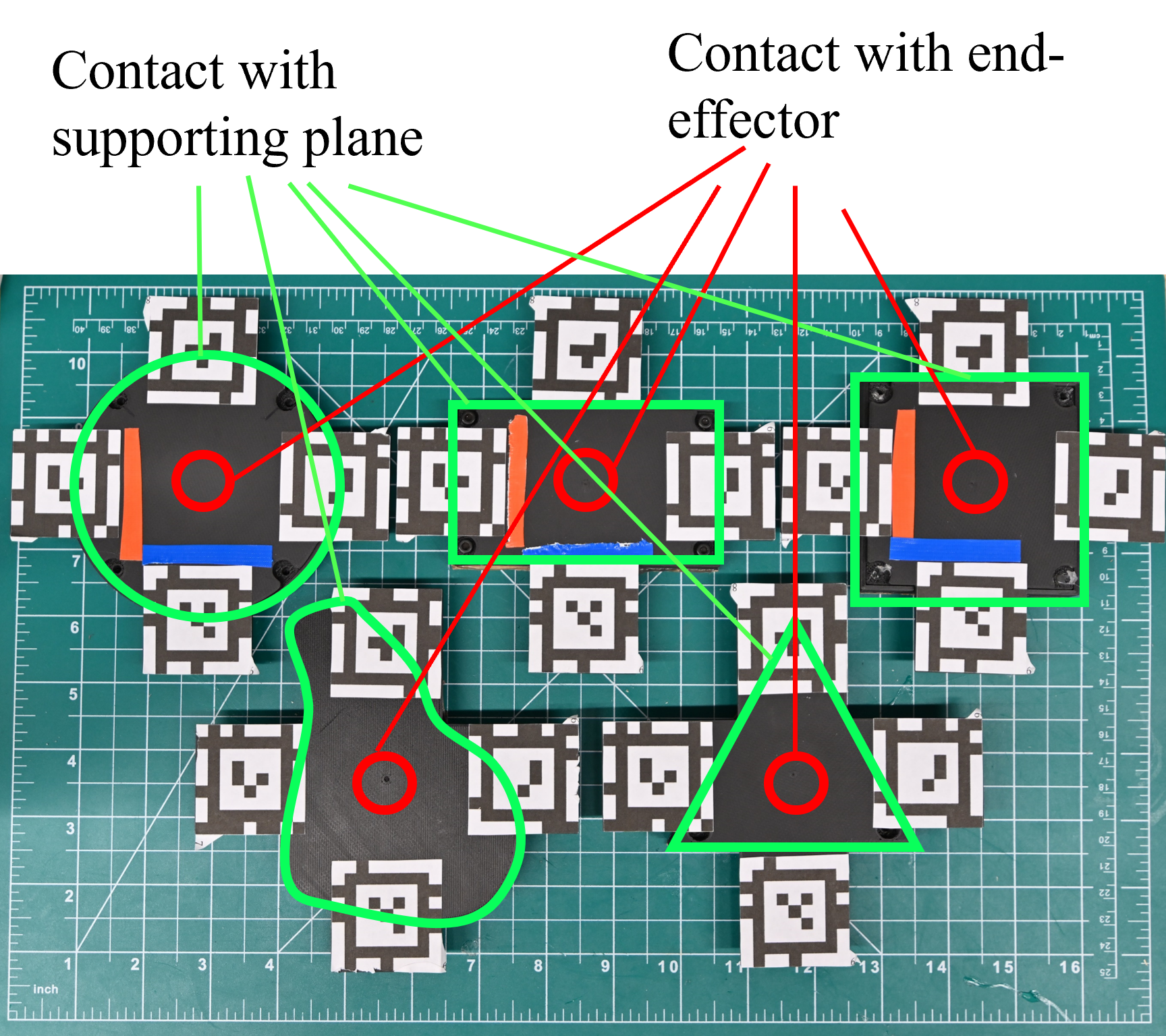}
    \caption{3D printed objects with 4 Apriltags to avoid occlusion while tracking. The support surfaces are: circle, rectangle, square, triangle and an irregular shape. They share the same top geometry that is center-aligned with the bottom surface geometric center. The relative pose of the Apriltags is common to all objects.}
    \label{fig:objects}
\end{figure}
The objects used for the task are shown in Fig. \ref{fig:objects}. 
They are 3D printed with carbon fiber filled nylon from ONYX. We chose five different shapes for the objects and their support surfaces: square, rectangle, round, triangle and an irregular shape. They all share the same material, thickness, mass, and relative position of Apriltags. An Intel Realsense D435 camera is fixed to the world frame for tracking. Four Apriltags are attached to the top of each object to prevent tracking failure due to occlusion, as shown in Fig. \ref{fig:objects}.  
The mass of each object is approximately 50 [g]. 

We used the Kuka MED LBR 14 R820 7 DOF robot to manipulate the end-effector. The robot is controlled in impedance Cartesian mode for compliance. Before executing each path, we set the normal force by lowering the end-effector until the force read by the F/T sensor reaches the desired value, then the path is executed at the same height, regardless of variation in normal force values. For planning and execution tasks, trajectories were computed using Matlab R2021b and sent to the robot through the ROS network. For the model validation, each path was executed 4 times with paths mirrored in translation and rotation to reduce potential sources of error due to calibration. 
\begin{figure*}
\centering
\includegraphics[width=16cm]{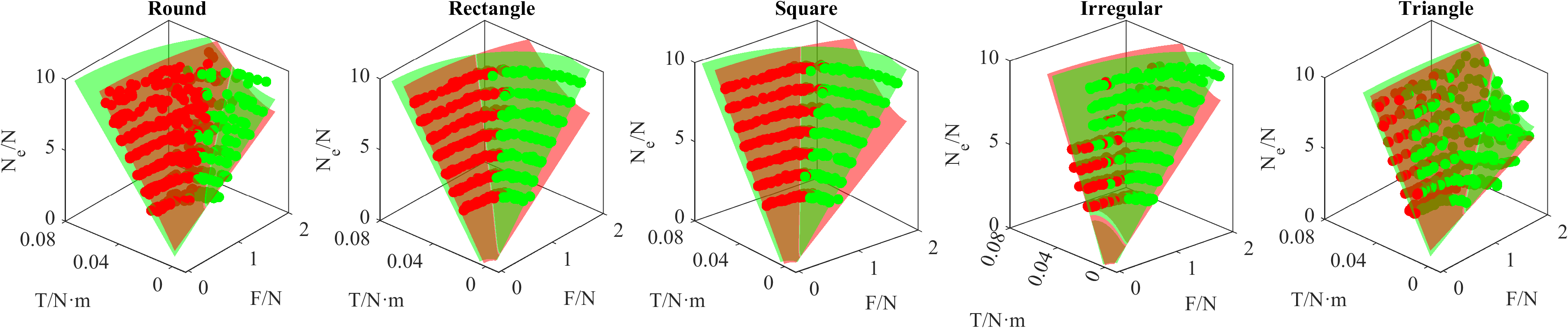}
\caption{\textbf{Model validation experiments:} Each point represents the wrench measured during a path segment. Red points denotes slippage, while green ones denotes no slippage. Overlayed are the fitted dual limit surface cones for each object. The red and green surfaces are fitted models.}
\label{fig:DLS_fitting}
\end{figure*}

\begin{figure*}
    \centering
    \includegraphics[width=0.7\textwidth]{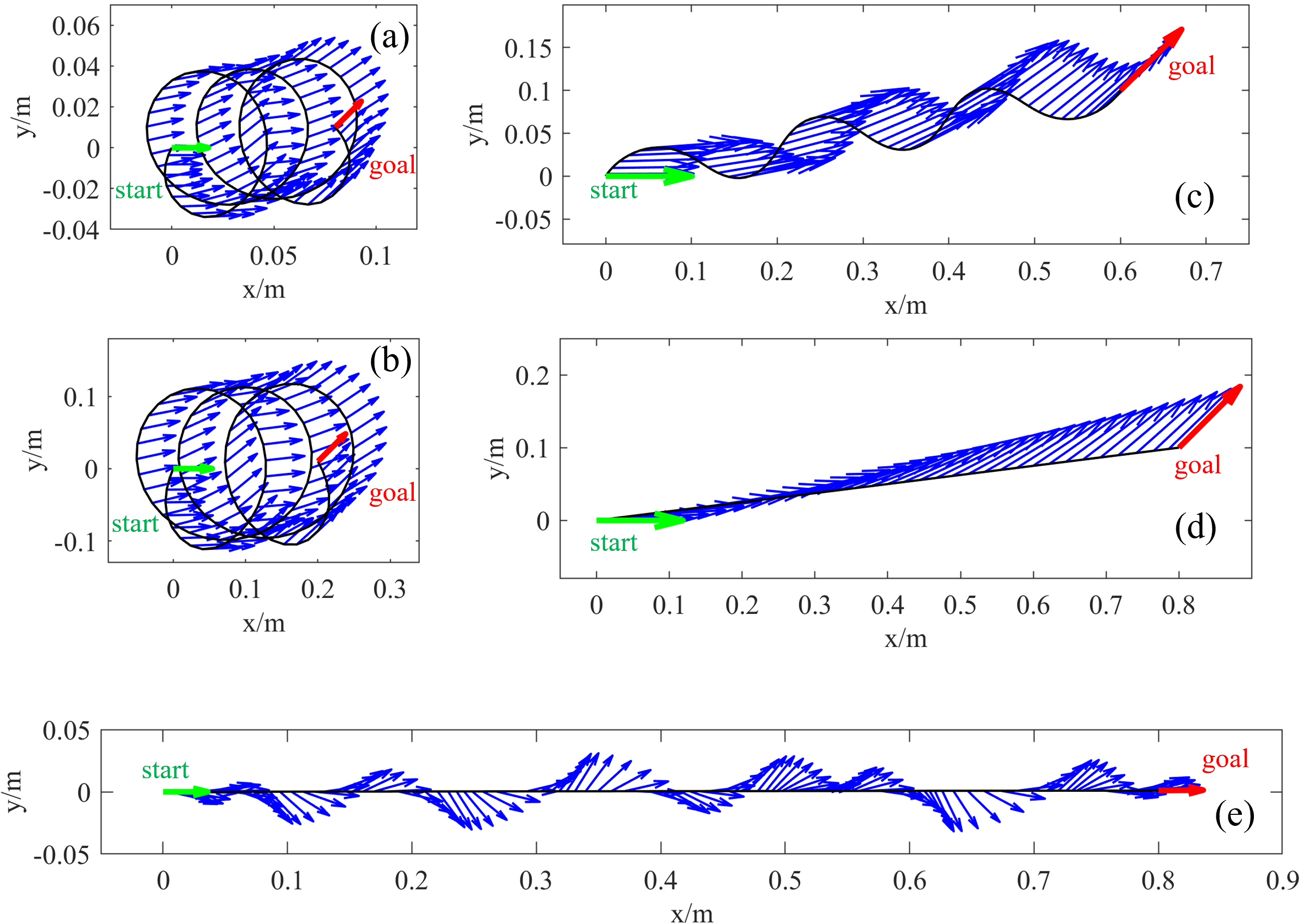}
    \caption{Exemplar planned paths for end-effector with different goal poses. Each subfigure shows the planned path using a black curve with the blue arrows representing the orientation along the path. Here, the green and red arrows represent the start and goal poses, respectively. Arrows are used to show both position and orientation at the same time. All paths start from the origin and end at given goal poses. 
    (a)-(d) show planned trajectories in cases (III) and (IV), (e) shows trajectory in case (II) and (V). The key insight is that the computed paths are often complex and winding to address the limited control authority in generating changes in orientation along the path. This limit is due to the frictional constraints between the end-effector and the object as well as the object and environment determined by the dual limit surface models.}
    \label{fig:planned_path}
\end{figure*}

\subsection{Model Identification and Evaluation} \label{subsection:model_validation}
In this section, we present experiments to evaluate the asymmetric dual limit surfaces and dual limit cones models. To generate data, we designed short path segments with the following features:
i) each segment has constant linear and angular velocity,
ii) the variation among segments provides a uniform distribution over the limit surface based on priori guesses of friction parameters. Each segment is repeated for normal forces in $N=\{3, ..., 9\} [\text{N}]$ .



To execute each segment, the robot end-effector moves above the object's center, then applies the normal force. Once the force read by the sensor is stable, the robot executes one of the designed segments. After executing each segment, the robot moves the object back to the center of the supporting plane with pure translation and repeats the previous steps. Each segment is executed 3 times.

For each segment, object and end-effector poses as well as force torque measurements are recorded: $\vec{d}=\{\vec{q}_e(0), \vec{q}_e(T), \vec{q}_o(0), \vec{q}_o(T), N_e, f_x, f_y, \tau \}$ where $T$ represents time at the end of execution. We used a data-set of 
$D=\{\vec{d}_1, ..., \vec{d}_{532}\}$ segments to fit the friction models by jointly optimizing over the parameters $\vec{\theta}=\{\mu_r, \mu_p, r_e, r_p\}$. We note that as part of the fitting, segments must be categorized as slipping and sticking. To evaluate whether the object has slipped w.r.t. to the end-effector during execution, the robot compares the difference between the initial and final poses of the object to the total change in its end-effector pose. If these differences exceed a threshold value, then the path is flagged for slippage. The threshold is set to be 0.05 [Rad] for orientation error, 0.005 [M] for position error.


Fig.~\ref{fig:DLS_fitting} shows the fitted models overlayed with the force torque data. Each point in Fig.~\ref{fig:DLS_fitting} represents the wrench values for each segment in the object frame. The green and red points represent slippage-free and slippage segments while the green and red surfaces are the fitted cone models, respectively. 
We note that when the end-effector slips w.r.t. the object, the assumptions that end-effector is always at the center of the object no longer holds, thus the red points are imprecise estimates of the dual limit cones shown in Fig. ~\ref{fig:dual_limit_surface_different_cases}.
This imprecision explains the uncertainty in the boundary between the two types of motion. We note that the results are presented in one quadrant for clarity as the data and models are axisymmetric due to the uniform pressure and homogeneous friction assumptions.



\subsection{Path Planning Evaluation}
{\renewcommand{\arraystretch}{1.3}
\begin{table*}
\begin{center}
\caption{Path execution RMSE results.}
\begin{tabular}{c c c c c c c}
\hline
\textbf{Object} & \textbf{Round} & \textbf{Square} & \textbf{Rectangle} & \textbf{Irregular} & \textbf{Triangle} & \textbf{Average}\\ \hline
Linear Planner Pos. RMSE [m] & 0.0013  & 0.0014 & 0.0013 & 0.0020 & 0.0019 & 0.0016 \\ 
Linear Planner Ori. RMSE [rad] & 0.1885 & 0.3354 & 0.1918 & 0.1646 & 0.1471 & 0.2055 \\ 
Proposed Planner Pos. RMSE [m] & 0.0009 & 0.0015 & 0.0013 & 0.0018 & 0.0019 & 0.0015 \\ 
Proposed Planner Ori. RMSE [rad] & 0.0071 & 0.0087 & 0.0072 & 0.0171 & 0.0176 & 0.0116 \\ \hline

\end{tabular}
\label{tab2}
\end{center}
\end{table*}
}

In this section, we evaluate the planner for cases (III) and (IV) over a variety of paths. For all paths, we use the parameters $(n, k_v, C_1, C_2) = (30, 1.25, 10, 1)$. 

\begin{figure*} 
\centering
\includegraphics[width=18cm]{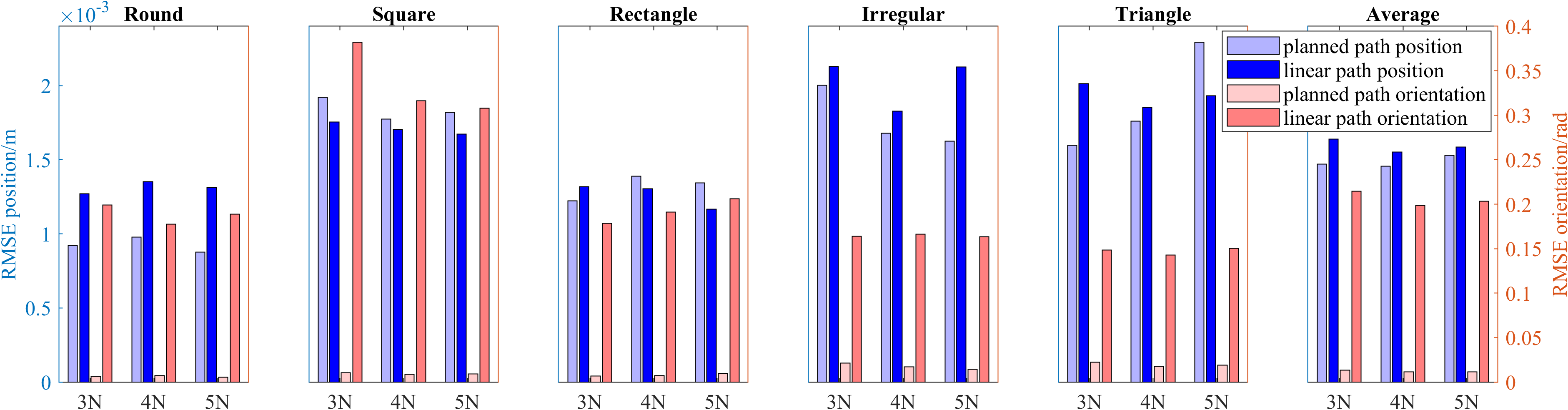}
\caption{RMSE of object pose with respect to the goal pose at the final step of the planner for both linear and our method for a variety of normal forces. We highlight that the orientation error of our methods planned paths drops to less than 10\% of linear paths. This is because our model explicitly considers frictional constraints and produces paths that enable the robot to maintain sticking contact to correctly orient the object. We note that linear position errors for both methods are approximately the same. Further, these position errors are negligible since the error due to Apriltag measurements has an RMSE of 0.0008 [m] in position and 0.002 [Rad] in orientation.}
\label{fig:path_error}
\end{figure*}

The step number $n=30$ balances the path smoothness and computational complexity. $k_v$ quantifies the amount of permissible rotation relative to translation, and the constants $C_1$ and $C_2$ balance the relative importance of smoothness and precision. 
Fig.~\ref{fig:planned_path} illustrates the object trajectory for several exemplar paths. The paths are shown as black curves and the green arrow and red arrows represent the start and goal poses. The blue arrows represent the pose (position and orientation) along the path. 

We note that as the ratio between the linear positional distance and the total rotation of the goal pose increases w.r.t. to the initial pose, the planned paths gradually change from a coil to a straight line. When the ratio is higher than \(1/k_v\), planned paths will all be the same as linear paths, as the linear motion satisfies the slippage-free constraints. 
But when the difference between the initial and goal orientation is small, the path planned is closer to a straight line, which fits our expectation. We also show one example of planning in case (II) and (V) in Fig. \ref{fig:planned_path}(e). Here, the object cannot translate without rotation, and the planner computes a straight translation with periodic rotations about the connecting line segment.

To evaluate the path planning algorithm, 
we measure the pose errors between the goal and achieved poses at the end of each path for both our proposed method and the linear planner. In all paths, we kept the end-effector moving in the same horizontal plane. The physical parameters are fit to cases (III) or (IV). We executed 162 paths with translation magnitudes ranging from 0.02 to 0.04 [m], rotation magnitudes from 0.5 to 0.9 [Rad], and normal forces from 3 to 5 [N]. These path parameters are chosen based on the size of supporting plane, size of the object, and friction coefficients. In addition to our approach, we consider a planner that linearly interpolates between the beginning and end poses in SE(2) and moves the end-effector along this path.

The RMSE of relative pose errors at the end of both paths are shown in Tab.~\ref{tab2} and in Fig.~\ref{fig:path_error}. The results show that using our proposed planner, there is a significant improvement in orientation error ($\approx 90\%$) when compared to the linear planner. This is because our model explicitly considers frictional constraints and produces paths that enable the robot to maintain sticking contact to correctly orient the object. We note that linear position errors for both methods are approximately the same. Further, these position errors are negligible since the error due to Apriltag measurements has an RMSE of 0.0008 [m] in position and 0.002 [Rad] in orientation. 

\begin{figure} 
    \centering
    \includegraphics[width=7.7cm]{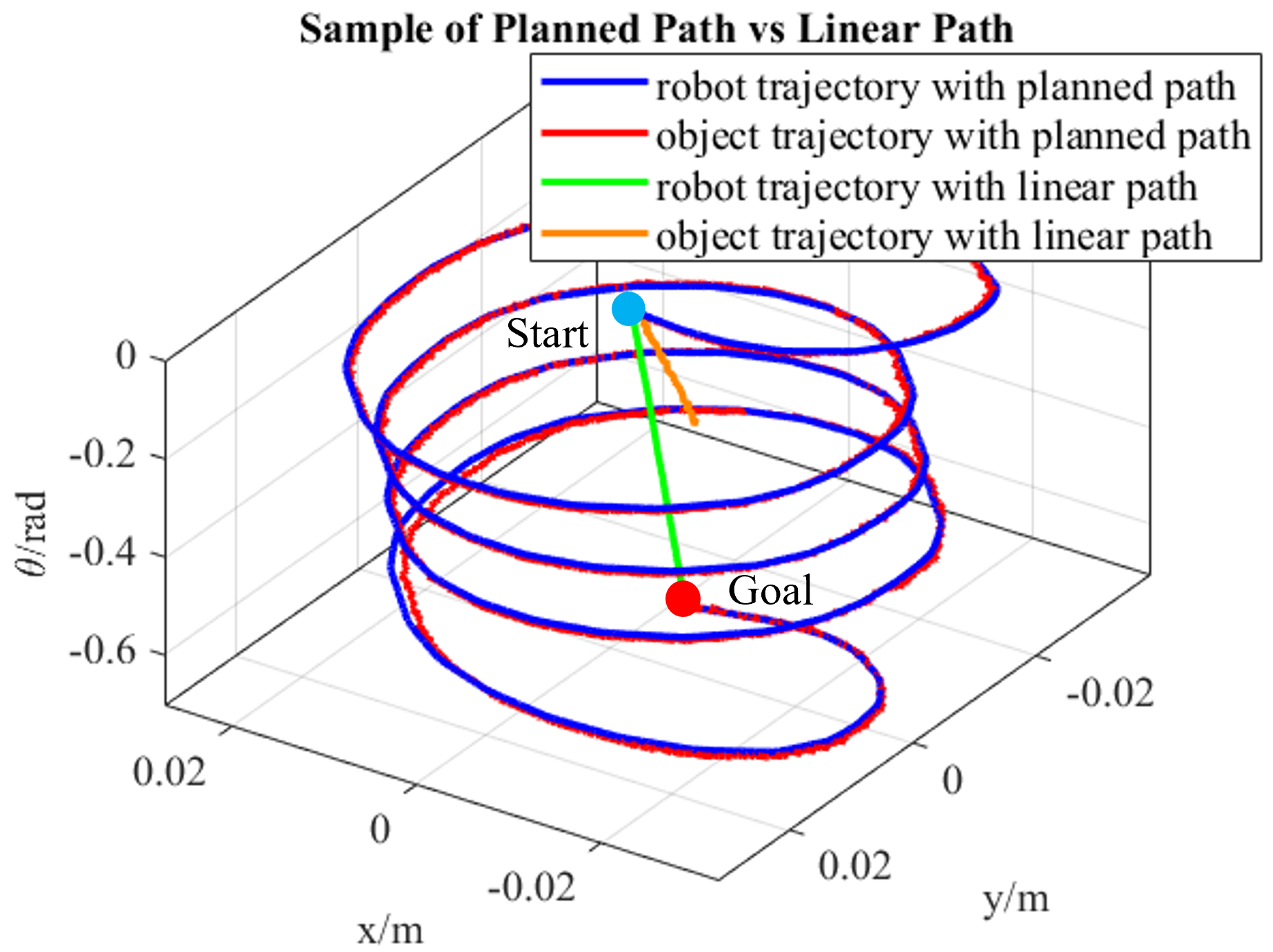}
    \caption{Visualization of a pair of exemplar paths in configuration space where the axes represent \(x\), \(y\) and \(\theta\). Both paths starts from the origin and end at [0, -0.01, -0.7]. The blue and red curves are the end-effector and object trajectories resulting from our method, while green and orange lines are end-effector and object trajectories with the linear path. 
    }
    \label{fig:path_sample_3d}
\end{figure}

\begin{figure} 
    \centering
    \includegraphics[width=9cm]{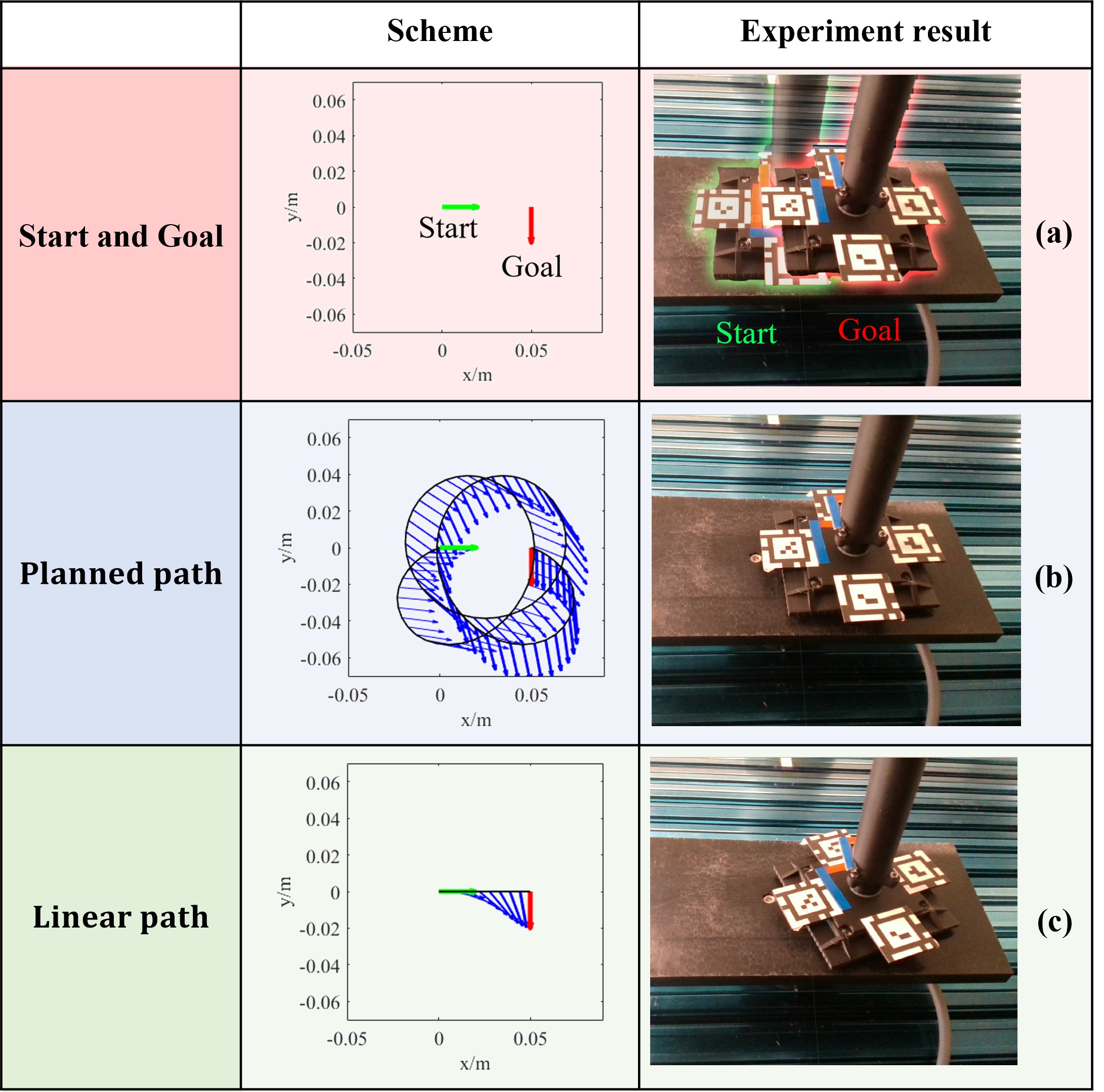}
    \caption{Visualization of the object path starting from the origin and ending at 5 [cm] to the right, with 90$^{\circ}$ clockwise orientation chage. In each row, the left shows the scheme of pose, the right shows the pose in real scene. (a) shows the start and desired goal pose, (b) shows the planned path and corresponding sliding result, (c) shows the linear path and corresponding sliding result.
    }
    \label{fig:path_sample_real_sim}
\end{figure}


Fig.~\ref{fig:path_sample_3d} shows example robot planned paths and corresponding object paths in configuration space for a given goal from both our approach and the linear planner. Here, the $(x,y)$ axes represent translations, and z axis represents rotation \(\theta\). The blue curve shows the end-effector path computed from our method and the red curve shows the corresponding realized object path. We note that the object closely follows the end-effector and reaches the desired goal. In contrast, the green curve is the linear end-effector path from the start pose to the goal, while the orange curve is the corresponding object trajectory. We highlight that the object fails to follow the end-effector with the linear path, where the deviation is especially pronounced in rotation. Fig.~\ref{fig:path_sample_real_sim} shows a visualization of paths and results for both scheme with their corresponding real-world experiments.

\section{Discussion \& Limitations}

In this paper, we developed a novel slippage-free path planning algorithm the allows the robot to precisely slide objects to arbitrary goal poses in the plane using only top contact. Our planner exploits a novel dual limit surface model that provides explicit slip/stick constraints and that are identified during task execution. 
We evaluated the model and planning algorithm with a Kuka iiwa Med robot with object of various geometries and demonstrated its effective performance over the naive linear path planner, particularly for significant reductions in rotation error. 

There are a few limitations with this work: First, while the dual limit surface experiments show that the model is accurate, the boundary of fitted surfaces does exhibit some uncertainty for points close to it. We hypothesize that the cause could be:
i) violations of the uniform pressure assumption,
ii) when the end-effector slips on top of the object, the two contact patches are no longer aligned,
iii) uncertainty in wrench measurement and force control that cannot be neglected.
Thus in the path planning and execution experiments we used more conservative parameters to guarantee the slippage-free path.
Fig. \ref{fig:path_error} shows that there's almost no difference in position error between the planned paths and naive paths. 
The reasons could be:
i) Apriltag and Realsense camera cannot provide enough accuracy,
ii) The deformation of the end-effector tip was ignored.

Future iterations of the planner and model can result in a more comprehensive planning algorithm that also takes the normal force into account, reducing the normal force when executing a more straight section along the path, so that the robot can apply less effort and decrease energy consumption. Also the model can be extended to compensate for the end-effector's possible deformation and address more general scenarios such as a tilting supporting surface and off center end-effector and object contact.

\bibliographystyle{plainnat}
\bibliography{references.bib}

\end{document}